%% file: iclr2026_conference.tex
\documentclass{article} %
\usepackage{iclr2026_conference,times}

\input{math_commands.tex}

\usepackage{hyperref}
\usepackage{url}

\usepackage{booktabs}
\usepackage{tabularx}
\usepackage[most]{tcolorbox}
\usepackage{xcolor}
\usepackage{soul} %
\usepackage{pifont}
\usepackage{subcaption}
\usepackage{wrapfig}
\usepackage{multirow}

\newcommand{\partitle}[1]{\smallskip \noindent \textbf{#1.}}
\newcommand{\hlgreen}[1]{\sethlcolor{green!20}\hl{#1}}
\newcommand{\hlred}[1]{\sethlcolor{red!20}\hl{#1}}
\newcommand{\hlblue}[1]{\sethlcolor{blue!20}\hl{#1}}

\definecolor{titlegray}{gray}{0.90}   %
\definecolor{contentgray}{gray}{0.98} %
\definecolor{darkgreen}{rgb}{0.0, 0.66, 0.0}
\newtcolorbox{promptbox}[1][]{
  enhanced,
  colframe=black,        %
  colback=contentgray,   %
  colbacktitle=titlegray,%
  coltitle=black,        %
  fonttitle=\bfseries,   %
  arc=8pt,               %
  outer arc=8pt,
  boxrule=0.6pt,         %
  titlerule=0.6pt,       %
  left=8pt, right=8pt, top=6pt, bottom=6pt, %
  title={#1}
}

\title{Direct Token Optimization: A Self-contained Approach to Large Language Model Unlearning}

\author{Hong kyu Lee \\
Department of Computer Science\\
Emory University\\
Atlanta GA, USA \\
\texttt{\{hong.kyu.lee\}@emory.edu} \\
\And
Ruixuan Liu \\
Department of Computer Science\\
Emory University\\
Atlanta GA, USA \\
\texttt{\{ruixuan.liu2\}@emory.edu} \\
\AND
Li Xiong \\
Department of Computer Science\\
Emory University\\
Atlanta GA, USA \\
\texttt{lxiong@emory.edu}
}

\iclrfinalcopy %
\begin{document}

\maketitle

\begin{abstract}
Machine unlearning is an emerging technique that removes the influence of a subset of training data (forget set) from a model without full retraining, with applications including privacy protection, content moderation, and model correction. The key challenge lies in ensuring that the model completely forgets the knowledge of the forget set without compromising its overall utility. Existing unlearning methods for large language models (LLMs) often utilize auxiliary language models, retain datasets, or even commercial AI services for effective unlearning and maintaining the model utility. However, dependence on these external resources is often impractical and could potentially introduce additional privacy risks. In this work, we propose direct token optimization (DTO), a novel self-contained unlearning approach for LLMs that directly optimizes the token level objectives and eliminates the need for external resources.
Given a sequence to unlearn, we identify two categories of tokens: target tokens, which capture critical knowledge for unlearning, and the remaining non-target tokens, which are crucial for maintaining the model utility. 
The former are used to optimize the unlearning objective, while the latter serve to preserve the model's performance.
The experimental results show that the proposed DTO achieves up to 16.8$\times$ improvement in forget quality on several benchmark datasets than the latest baselines while maintaining a comparable level of model utility.

\end{abstract}

\section{Introduction}

Machine unlearning aims to remove the effect of a subset of training data (referred to as the forget set) from a trained model~\citep{cao2015towards}. The concept was introduced in response to data protection regulations such as General Data Protection Regulation (GDPR)~\citep{mantelero_eu_2013}, which established the ‘right to be forgotten’. Beyond privacy considerations, unlearning has also become important for removing copyrighted material, unsafe or harmful content inadvertently incorporated during training~\citep{liu2024towards}. A successfully unlearned model should fully eliminate the influence of forget set (unlearning efficacy), and preserve overall performance (model utility). Additionally, the unlearning algorithm should be more efficient than retraining (efficiency).

Large language models (LLMs) have demonstrated impressive performance across various tasks~\citep{chen2024survey,xiao2025comprehensive}, and their tendency to strongly memorize training data~\citep{carlini2022quantifying,tirumala2022memorization} makes unlearning both urgent and challenging.
Typically, LLMs are pre-trained with a large general corpus, and then fine-tuned with a smaller and task-specific fine-tuning dataset for downstream tasks~\citep{ziegler2019fine}.
Existing works have demonstrated that both pre-trained and fine-tuned model are susceptible to memorize sample-specific content~\citep{wang2024unique,fu2024membership}.
Moreover, fine-tuning data is more prone to memorization due to its domain-specific distribution diverging from the general knowledge~\citep{zeng2023exploring,akkus2025generated}. Our work aim to unlearn fine-tuning data from fine-tuned LLMs, providing a practical and generalizable solution for both privacy protection and content control.  

\begin{figure}[h]
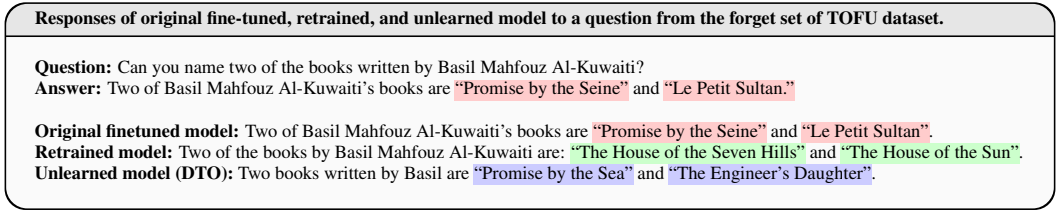

\scriptsize
\begin{promptbox}[Responses of original fine-tuned, retrained, and unlearned model to a question from the forget set of TOFU dataset.]
\textbf{Question:} Can you name two of the books written by Basil Mahfouz Al-Kuwaiti? \\
\textbf{Answer:} Two of Basil Mahfouz Al-Kuwaiti's books are \hlred{``Promise by the Seine''} and \hlred{``Le Petit Sultan.''} \\

\textbf{Original finetuned model:} Two of Basil Mahfouz Al-Kuwaiti's books are \hlred{``Promise by the Seine''} and \hlred{``Le Petit Sultan''}. \\
\textbf{Retrained model:} Two of the books by Basil Mahfouz Al-Kuwaiti are: \hlgreen{``The House of the Seven Hills''} and \hlgreen{``The House of the Sun''}.\\
\textbf{Unlearned model (DTO):} Two books written by Basil are \hlblue{``Promise by the Sea''} and \hlblue{``The Engineer's Daughter''}.
\end{promptbox}
\caption{Examples of Finetuned LLM Unlearning}
\label{fig:responses}
\end{figure}

We illustrate the effect of unlearning on a fine-tuned model using the TOFU dataset~\citep{maini2024tofu}, which contains synthetic author profiles with corresponding question-answer (QA) pairs. For questions concerning authors in the forget set (targeted for unlearning), we compare responses generated by the original fine-tuned model, retrained model (excluding the forget set), and the unlearned model using our approach in  Figure~\ref{fig:responses}. 
The fine-tuned model generates the exact original answer, since the QA pair  is included in the fine-tuning data. The retrained model, trained on the fine-tuning data excluding the forget set, serves as a gold-standard for unlearning. As it has not seen the QA pair, it provides an alternative response (highlighted in green). Notably, the initial portion of the responses from both models are almost identical, reflecting the general linguistic capability of the model rather than memorization on training data. This highlights two key objectives for fine-tuned LLM unlearning: (1) preserve the model's linguistic capability, and (2) remove the core knowledge of the forget set so that the model does not generate specific words encoding that knowledge (highlighted in green). 
The response from the unlearned model produced by our method (DTO) demonstrates these desired behaviors.  It preserves linguistic fluency while effectively unlearning the targeted knowledge. As a result, it provides an alternative response (highlighted in blue). %

\begin{table}[h]
    \centering
    \scriptsize
    \begin{tabularx}{\textwidth}{p{0.35\textwidth}XXXX|X}
        \toprule
        Title & Original model  & Retain data & Auxiliary Model & LLM Service & Practicality \\
        \midrule
        DPO~\citep{rafailov2023direct} & \textcolor{darkgreen}{\ding{52}}  & \textcolor{darkgreen}{\ding{56}} & \textcolor{darkgreen}{\ding{56}} & \textcolor{darkgreen}{\ding{56}} & \textcolor{darkgreen}{\ding{52}} \\
        NPO~\citep{zhang2024negative} & \textcolor{darkgreen}{\ding{52}}  & \textcolor{darkgreen}{\ding{56}} & \textcolor{darkgreen}{\ding{56}} & \textcolor{darkgreen}{\ding{56}} & \textcolor{darkgreen}{\ding{52}} \\
        WHP~\citep{eldan2023s} & \ding{56}  & \textcolor{red}{\ding{52}} & \textcolor{red}{\ding{52}} & \textcolor{red}{\ding{52}} & \textcolor{red}{\ding{56}} \\
        LLMU~\citep{yao2024machine} & \ding{52} & \textcolor{red}{\ding{52}} & \ding{56} & \ding{56} & \textcolor{red}{\ding{56}} \\
        ECO-Prompts~\citep{liu2024large} & \ding{56} & \ding{56} & \textcolor{red}{\ding{52}} & \ding{56} & \textcolor{red}{\ding{56}} \\
        ULMR~\citep{shi2024ulmr} & \ding{52}  & \textcolor{red}{\ding{52}} & \textcolor{red}{\ding{52}} & \ding{56} & \textcolor{red}{\ding{56}}\\
        FLAT~\citep{wang2024llm} & \textcolor{darkgreen}{\ding{56}} & \textcolor{darkgreen}{\ding{56}} & \textcolor{darkgreen}{\ding{56}} & \textcolor{darkgreen}{\ding{56}} & \textcolor{darkgreen}{\ding{52}} \\ 
        TPO~\citep{zhou2025not} & \ding{56} & \ding{56} & \ding{56} & \textcolor{red}{\ding{52}}  & \textcolor{red}{\ding{56}}\\
        TSFD~\citep{kumar2025selective} & \ding{52} & \ding{56} & \textcolor{red}{\ding{52}} & \ding{56} & \textcolor{red}{\ding{56}} \\
        \midrule
        DTO(Ours) & \textcolor{darkgreen}{\ding{52}} & \textcolor{darkgreen}{\ding{56}} & \textcolor{darkgreen}{\ding{56}} & \textcolor{darkgreen}{\ding{56}} & \textcolor{darkgreen}{\ding{52}} \\
        \bottomrule
    \end{tabularx}
    \caption{List of the most recent LLM unlearning frameworks and their assumptions. The check-mark (\ding{52}) and cross (\ding{56}) denote that the framework requires the resource or not.}
    \label{tab:ref_model}
\end{table}

While several LLM unlearning algorithms have been proposed to unlearn fine-tuning data from a fine-tuned model, their underlying dependence on external resources limits their practicality in real-world scenarios. Table~\ref{tab:ref_model} summarizes the external resources that existing LLM unlearning frameworks depends on, and their overall practicalities. A red check mark indicates that the use of the corresponding resource may be impractical. %
Some works rely on retain set, consisting of the fine-tuning data excluding the forget set, to maintain model utility~\citep{yao2024machine, wang2025rethinking}. 
However, assuming access to the retain set may be unrealistic, as data regulations such as GDPR impose strict limitations on storing and reusing raw data~\citep{basaran2025a}. Other works assume availability of auxiliary models. These include (1) prompt classifiers~\citep{liu2024large,deng2025guard}, which detect inputs related to the forget data during inference, (2) additional LLMs~\citep{eldan2023s,kumar2025selective} obtained by further fine-tuning on the forget set.
The auxiliary models %
might not be available due to the high computational and storage costs of maintaining additional LLMs %
as well as potential privacy risk that the knowledge of forget set being embedded and persisting in the auxiliary LLMs.
Finally, some works leverage existing LLM services. \citet{eldan2023s,shi2024ulmr,zhou2025not} utilized ChatGPT-4 to generate custom dataset from the raw forget set. However, external AI services are generally untrusted as they may collect input data for future training~\citep{openai_privacy_policy_2025}. Exposing sensitive forget set directly to such services could cause additional privacy risk~\citep{wu2024unveiling}.
A few works do not require external sources, however they suffer from poor unlearning efficacy or model utility. 

\partitle{Contributions} To address these issues, we propose Direct Token Optimization (DTO) for unlearning fine-tuned LLMs, which does not rely on any external resource, such as auxiliary models, retain dataset, or %
external AI services. Given a token sequence from forget set, DTO identifies the target tokens that are most critical for representing the knowledge of the sequence and utilize them for unlearning. 
Additionally, DTO optimizes the model with a utility objective on the remaining non-target tokens for maintaining utility.

Different from existing works that rely on human annotator~\citep{yang2025exploring} or external LLMs such as ChatGPT~\citep{zhou2025not} to select tokens for unlearning, we propose Delta-score, an assistance-free token selection strategy for LLM unlearning inspired by a study of sequence memorization~\citep{stoehr2024localizing}. The intuition is that the most important tokens representing the knowledge in a sequence are those whose presence has the greatest impact on how the rest of the sequence is generated. 
Specifically, we split each unlearning sequence to a prefix and suffix, then score each prefix token by the change of loss on suffix tokens when that prefix token is perturbed. 
Then the prefix tokens with highest scores - those with the greatest influence and thus encoding the core knowledge - are chosen as target tokens, on which DTO conducts gradient ascent for unlearning. 
For the remaining non-target tokens, DTO minimizes the KL-divergence between the logits of the updated and original model to maintain model utility. This ensures that the model forgets the targeted knowledge while preserving its general language ability.

To demonstrate the efficacy of DTO, we conduct experiments with various LLMs on TOFU~\citep{maini2024tofu} and MUSE~\citep{shi2024muse} benchmarks. 
We compare DTO to several baselines, including the most recent method FLAT~\citep{wang2024llm}, under the same assumptions of relying on only the original model and forget set. %
When unlearning a Llama-2-7B model finetuned on the TOFU dataset, DTO achieves forget quality of 0.918, a significant improvement over FLAT~\citep{wang2024llm} (0.054). We summarize our contributions as follows.

\begin{enumerate}
    \item We propose Direct Token Optimization (DTO), a self-contained approach for LLM unlearning that does not require any auxiliary model, retain dataset or external AI services. DTO selects target tokens for unlearning and remaining non-target tokens for utility preservation, and optimizes them accordingly.  
    \item Inspired by previous studies on memorization, DTO proposes the delta-score for selecting target tokens and non-target tokens, and conducts gradient ascent on target tokens to achieve unlearning while regularizing on non-target tokens to maintain model utility. 
    \item We conduct experiments and compare the results with state-of-the-art LLM unlearning methods under the same assumptions. The results show that DTO improves the forget quality substantially with only minimal utility degradation.
\end{enumerate}

\section{Related Work}
Machine unlearning was first introduced by~\citet{cao2015towards} and has been extensively studied for classification models~\citep{kurmanji2023towards,tarun2023fast,cha2024learning,huang2024unified}. 
This field encompasses two primary approaches.
\textbf{Exact unlearning}~\citet{bourtoule2021machine} completely removes the influence of target samples by data partitioning and retraining but often at significant computational cost. \textbf{Approximate unlearning} methods iteratively update model parameters so that the unlearned model's behavior approaches that of a retrained model. Notable approaches include maximizing KL-divergence over target sample logits~\citep{kurmanji2023towards,huang2024unified}, injecting calibrated noise~\citep{tarun2023fast}, optimizing the embedding space~\citep{lee2024contrastive}, and leveraging adversarial examples~\citep{ebrahimpournot}. Certified unlearning refers to approximate unlearning methods that come with certifiable unlearning guarantees~\citep{guo2019certified,koloskova2025certified}.

Unlearning LLMs is more challenging compared to unlearning classification models. The label space of classification model is usually small and fixed, enabling unlearning through the disassociation of 
labels of forget set~\citep{kurmanji2023towards}. The unlearning efficacy and model utility can be directly evaluated by accuracy on forget set and test data~\citep{lee2024contrastive}. 
In contrast, LLMs generate sequences from a vast and unbounded text space, which fundamentally complicates the unlearning process. 
Unlike classification tasks, unlearning in LLMs cannot be achieved by simple word suppression~\citep{cooper2024machine}. Instead, effective unlearning method should minimize the generation of certain sequences or facts~\citep{eldan2023s,yao2024machine}.

LLM unlearning can be broadly categorized into two types: unlearning pre-trained models and unlearning fine-tuned models.
While several methods for unlearning pre-trained models and related benchmarks have been proposed~\citep{li2024wmdp,jin2024rwku,liu2024revisiting}, lack of access to the original pre-training datasets~\citep{yao2024machine} prevents accurate ground truth evaluation and makes scaling to real-world scenarios challenging~\citep{zhou2024limitations}.
Unlearning fine-tuned LLMs defines a forget set which was a part of the fine-tuning dataset, and most works are evaluated on benchmarks such as TOFU~\citep{maini2024tofu} and MUSE~\citep{shi2024muse}.

Most works unlearn fine-tuned LLMs by approximate unlearning,%
such as preference optimization~\citep{zhang2024negative,fan2024simplicity}, second-order update~\citep{jia2024soul,gu2024second} and instruction fine-tuning~\citep{shi2024ulmr}. 
As shown in Table \ref{tab:ref_model}, most of them utilize retain set or auxiliary models for better unlearn efficacy and model utility~\citep{yuan2024closer,wang2025rethinking, krishnan2025not}.
However, the availability of retain set or a surrogate dataset with the same distribution~
\cite{basaran2025a} can not be guaranteed
due to privacy regulations, or practical resource limitations~\citep{chundawat2023zero}. 
Some works use auxiliary LLM models, obtained by fine-tuning a pre-trained model with the forget set~\citep{wang2025rethinking,ji2024reversing} or further finetuning the finetuned model~\citep{eldan2023s} with the forget set. %
However, the knowledge of forget data still remains and the continue fine-tuning introduces extra training cost besides unlearning.
Optimization-free methods rely on prompt classifiers to detect forget set inputs and subsequently activate LoRA adapters~\citep{gao2024large,deng2025guard} or corrupt the embeddings~\citep{liu2024large} to prevent the model from answering the target knowledge. These methods often suffer from limited forget quality, dependency on detection accuracy, and the impracticality of adding prompt classifiers due to training, scalability, and deployment challenges. 
A more realistic setting is conducting unlearning only with the original model and the forget set. This includes the direct preference optimization (DPO)~\citep{rafailov2023direct} and negative preference optimization (NPO)~\citep{zhang2024negative}. The most recent is FLAT~\citep{wang2024llm}, which uses $f$-divergence to steer parameters towards generating refusal responses. While the framework well preserves the model utility, their forget quality shows a clear limitation. 

\section{Direct Token Optimization}

\partitle{Problem Definition}
Given a forget set $\mathcal{D}_F$, retain dataset $\mathcal{D}_R$, and an original LLM $\theta_o$ fine-tuned with $\mathcal{D}_F \cup \mathcal{D}_R$ from a pre-trained LLM $\theta_p$, LLM unlearning aims to produce an unlearned model $\theta_u$ that approximates a hypothetical model $\theta_{rt}$ that was fine-tuned only with $D_R$.
We assume the unlearner has access only to the forget set $\mathcal{D}_U$ and the original model $\theta_o$, without access to other auxiliary models or retain dataset $D_R$. 
This setting follows DPO~\citep{rafailov2023direct}, NPO~\citep{zhang2024negative} and FLAT~\citet{wang2024llm}.

\partitle{Intuition} As shown in Figure~\ref{fig:responses}, the semantic differences between responses from the fine-tuned and retrained model arise primarily from the words highlighted in green and red. This suggests that a small set of tokens are crucial for conveying dataset-specific knowledge. Motivated by this, we aim to unlearn by suppressing the model's ability to generate those crucial tokens while preserving the overall sentence structure that is useful for other queries. We call the crucial tokens as target tokens and the rest as non-target tokens. We propose the delta score to identify target tokens in forget set.

\partitle{Delta Score: Identifying Target Tokens} 
Prior works~\citep{yang2025exploring} find that 
performing unlearning  on unique identifier words, such as names and locations, are effective for unlearning samples or entities. 
However, focusing only on these identifiers is often insufficient, as models may also memorize surrounding context or other tokens that encode the same knowledge. 
Instead of relying on linguistic rules to pick these identifiers, 
we adopt a more general token-level perspective, naturally aligning with how LLMs process and memorize unique tokens during fine-tuning~\citep{huang2024demystifying}. 
This perspective allows our method to target key tokens that contribute to the memorized knowledge, ensuring more effective unlearning.
Intuitively, tokens with low per-token-loss indicate stronger memorization and are natural candidates for target tokens as they contribute more to memorizing the sample. However, linguistically important yet semantically uninformative tokens also have low per-token loss, due to the frequent exposure during the pre-training stage~\citep{duan2024membership}. Unlearning these causes a detrimental effect on the linguistic fluency of the model. Thus, it is challenging to distinguish target tokens for unlearning from the linguistically important tokens. Most recent works  rely on ChatGPT~\citep{zhou2025not} and human annotators~\citep{yang2025exploring} to identify these tokens, but both approaches are impractical as external AI assistance can be unreliable and untrusted, and human annotation is costly and inconsistent, and both can severely affect the unlearning performance.

Instead, we identify target tokens as the tokens in the prefix that are critical for eliciting the model's memorized output. This is motivated by a study that analyzed memorization in LLMs through perturbation~\citep{stoehr2024localizing}. The study fine-tuned a GPT-Neo model with the PILE dataset~\citep{gao2020pile}, where each sample is a 100-token paragraph. To quantify memorization, the authors provided the model with a prefix consisting of the first 50 tokens and generated the remainder with greedy decoding; they then measured (1) average NLL over generated sequences and (2) number of generated tokens exactly matching the ground-truth suffix (last 50 tokens of the paragraph). By perturbing one token from the prefix at a time, they identified which token perturbation produced the largest response difference and the biggest NLL spike. Their analysis highlighted that perturbing specific token in the prefix can introduce a significant change in the output, indicating that certain tokens in a prefix serve as a trigger for generating the memorized suffix.

We extend these findings for unlearning purposes. 
Our insight is that not all tokens in the forget set contribute equally to the model's memorized knowledge; therefore, suppressing this memorized knowledge from the forget set can be most effectively achieved by suppressing the prefix triggers identified by the perturbation analysis.
Once these target tokens are identified, taking the gradient ascent on them reduces likelihood of generating themselves, naturally leads to reducing the generation of the memorized knowledge.

While \citet{stoehr2024localizing} measures NLL over the generated response conditioned on the prefix and partially generated response, our delta-score computes NLL for each suffix token conditioned on the prefix and the original suffix. This gives a stronger signal.  When NLL is obtained through conditioning on the {\em generated} sequence, it naturally decreases toward the end, regardless of whether the response matches the original suffix or not. In contrast, delta-score amplifies the NLL loss by forcing the model to condition on both perturbed prefix and the original suffix. This accurately captures the models' disagreement over the original suffix.

Let $\mathcal{D}_F = \{s^{i}\}^N_{i=1}$ be a forget set with $N$ samples. Let $s^{i} = \{x^i_1, \cdots x^i_t \cdots x^i_{T_i}\} \in \mathcal{V}^{T_i}$ be a sequence of tokens $x^i_t$ from the vocabulary $\mathcal{V}$ with the length $T_i$. 
Let a pivot $1 \leq q_i < T_i$ divide $s^i$ into a prefix $\{x^i_1 \cdots x^i_{q_i}\}$ and a suffix $\{x^i_{q_i+1} \cdots x^i_{T_i}\}$. Let a subsequence with size $t-1$ with perturbation on its $r$-th token as $\tilde{x}^{i}_{<t}:=\left(x^{i}_{1},\ldots,x^{i}_{r-1},\,\tilde{x}^{i}_{r},\,x^{i}_{r+1},\ldots,x^{i}_{t-1}\right)$. We define the delta score $\Delta^i_r$ at position $r \leq q_i$ of sequence $s^i$ as follows, and the perturbed token is randomly selected from special tokens (`UNK', `\#', etc.).
\begin{align}
\Delta^i_r = \sum^{T_i}_{t \ = \ q_i+1}\bigg(\log p_{\theta}\left(x^{i}_{t} \,\big|\,x^i_{<t}\right)\bigg) -\sum^{T_i}_{t \ = \ q_i+1} \bigg(\log p_{\theta}\left({x}^{i}_{t} \,\big|\,\tilde{x}^i_{<t}\right)\bigg) 
\label{eq:delta}
\end{align}
Equation~\ref{eq:delta} defines the delta-score as the difference between the average NLL over all suffix tokens when $r$-th token in prefix is present and when it is perturbed. 
For each unlearning sequence, we select top-$k\%$ highest scoring tokens as target tokens $\mathcal{T}^i_{k\%} = \{x^i_r \ \vert \ r \in \textrm{Top-}k\%\left(\Delta^i_r\right)\}$, and set the rest as non-target tokens $\mathcal{N}^i = s^i \setminus {T}^i_{k\%}$.

\partitle{Unlearning Using Target Tokens} Target tokens are used for unlearning knowledge from each sequence. We conduct gradient ascent using the loss of predicting the target tokens. 
\begin{align}
\theta_u \leftarrow \theta_u + \eta \nabla_{\theta_u} \sum_{x_t \in \mathcal{T}^i_{k\%}}{\log p_\theta\left(x_t \ \vert \ x_{<t}\right)},
\label{eq:ga}
\end{align}
where $\eta$ is a step size. Non-target tokens are used for maintaining linguistic fluency and general model utility. For each sequence, we minimize the KL-divergence between the logits of these tokens from the original model $\theta_o$ and the corresponding logits from $\theta_u$.
\begin{align}
    \theta \leftarrow \theta - \nabla_\theta \sum_{x_t \in \mathcal{N}^i}{\textrm{KL}\big( f_{\theta_0}\left(x_{<t}\right) \ \vert \ f_{\theta_u}\left(x_{<t}\right)\big)}
\label{eq:kl}
\end{align}
where $f_\theta$ is the model output logit after softmax.
While the default DTO is designed to update the model using the non-target tokens, we perform an ablation study in the experiments to compare it with a version of DTO without the KL update.
To avoid potential gradient conflicts, each unlearning step (\ref{eq:ga}) and minimizing KL-divergence step (\ref{eq:kl}) are performed in an alternating manner. We orthogonalize one gradient with respect to the other, which further reduces gradient conflicts and improve both unlearn efficacy and model's utility~\citep{kodge2024deep} Refer to Appendix~\ref{app:grad} for more details.

\section{Experiments}
\subsection{Experimental Setup}
\partitle{Datasets \& Evaluation Metrics}
We evaluate our method on two LLM unlearning benchmarks.
The TOFU~\citep{maini2024tofu} dataset has 4,000 question and answer pairs of fictitious authors for finetuning any LLMs. It provides multiple tasks of unlearning 1\%, 5\% and 10\% of the training dataset. %
The dataset provides following evaluation metrics: \textbf{Model utility}, obtained from the aggregated score of Rouge-L, normalized probability over answers, and truth ratio (probability of correct answer over the incorrect answer) on question and answer pairs of remaining, real authors and real world dataset; and \textbf{Forget quality}, measured using the Kolmogorov-Smirnov test on the truth ratios using both unlearned and retrained model. A successfully unlearned model should exceed 0.05~\cite{mekala2024alternate}. We also show the Forget-Rouge as a proxy to evaluate the memorization of the model.
We use finetuned version provided by authors for Llama 3.2-1B and Llama 2-7B model~\citep{touvron2023llama}.
For MUSE benchmark~\citep{shi2024muse}, we use the MUSE-book, which consists of Chapter 2 of the Harry Potter series. Muse benchmark offers following evaluation metrics: \textbf{Verbatim Memorization (VerbMem)}, obtained via Rouge-L F1 scores, shows the exact sequence memorization from the model; \textbf{Knowledge Memorization (KnowMem)}, obtained via Rouge-L scores on question and answer evaluation dataset, evaluates how the model retains factual knowledge of the unlearning contents; and \textbf{Privacy Leakage (PrivLeak)}, obtained via membership inference attack~\citep{carlini2022membership}, evaluates if the model's response after unlearning still reveals that the forget set was part of their fine-tuning data. It is a normalized AUC difference of the membership inference attack on the unlearned model and the retrained model. The negative value means the model is under-unlearned, and positive means over-unlearned. The ideal model should have the value close to zero.

\partitle{Baselines} We compare our framework with baselines that have the same assumption (access to forget set and original model only). NPO~\citep{zhang2024negative} unlearns by conducting preference optimization to reject answering questions in the forget set. DPO~\citep{rafailov2023direct} unlearns by up-weighting generation of a rejection template for the forget set. FLAT~\citep{wang2024llm} is the most recent and state-of-the-art framework that steers the model to generate rejection template over the original answer by maximizing $f$-divergence\footnote{The official implementation was inaccessible, hence we re-implemented this baseline, and confirmed that the result is comparable. Refer to Appendix~\ref{app:flat} for details.}. As proposed from the paper, we use Kullback–Leibler (KL), Total Variation (TV), Jensen–Shannon (JS) and Pearson (P) divergences. We search hyper-parameters to find the best model utility and forget quality tradeoff for each baseline.

While not directly comparable, we also include LLMU~\citep{yao2024large} which requires retain dataset.  It conducts gradient ascent on all responses and minimizes the KL-divergence between the original model and the unlearned model over the retain set for model utility.  %
In addition, we compare our token selection strategy and the unlearning result with Token Preference Optimization (TPO)~\citet{zhou2025not} which uses ChatGPT to identify target words to unlearn.

\subsection{Experimental Results}

\begin{table*}[t] %
    \centering
    \scriptsize
    \begin{tabularx}{\textwidth}{l*{7}{>{\centering\arraybackslash}X}}
        \toprule
        &
        \multicolumn{3}{c}{\textbf{Llama2-7B}} & \multicolumn{3}{c}{\textbf{Llama3.2-1B}} \\
        \cmidrule(lr){2-4} \cmidrule(lr){5-7}
         & Forget Quality ($\uparrow$) & Model Utility ($\uparrow$) & Forget Rouge-L\,($\downarrow$) & Forget Quality ($\uparrow$) & Model Utility ($\uparrow$) & Forget Rouge-L\,($\downarrow$)\\
        \midrule
        Original LLM  & 2.183e-06 & 0.6346 & 0.8849 & 3.383e-06 & 0.6218 & 0.8168 \\
        Retrained LLM  & 1.0000      & 0.6267 & 0.4080 & 1.0000 & 0.6168 & 0.4045 \\
        \midrule
        LLMU          & 0.0541 & 0.6225 & 0.4472 & 0.0301 & 0.5876 & 0.4671 \\
        DPO           & 0.0541 & 0.6219 & 0.5724 & 0.0541 & 0.5191 & 0.4752 \\
        NPO           & 0.0068 & 0.6242 & 0.4523 & 0.2650 & 0.5608 & 0.2447\\
        \midrule
        FLAT (TV)     & 0.0541 & 0.6199 & 0.4366 & 0.1649 & 0.5687 & 0.2543 \\
        FLAT (KL)     & 0.0301 & 0.6393 & 0.4971 & 0.1430 & 0.5639 & 0.2688 \\
        FLAT (JS)     & 0.0970 & 0.6214 & 0.4252 & 0.0286 & 0.5548 & 0.3793 \\
        FLAT (P)      & 0.0541 & 0.6239 & 0.4523 & 0.1649 & 0.5578 & 0.2567\\
        \midrule
        \textbf{DTO w/o KL (Ours)} & \textbf{0.9188} & 0.5948 & \textbf{0.3725} & \textbf{0.9188} & 0.5375 & 0.2893 \\
        \textbf{DTO (Ours)} & \textbf{0.7659} & 0.6002 & 0.3978 & \textbf{0.7659}& 0.5529 & \textbf{0.2382} \\  
        \bottomrule
    \end{tabularx}
    \caption{Experimental results on TOFU 1\% dataset.}
    \label{tab:tofu01}
\end{table*}

\partitle{Unlearn Efficacy and Model Utility}
Table~\ref{tab:tofu01} shows the result of unlearning 1\% of the TOFU dataset with DTO and baselines. We used top $k=20$\% for selecting target tokens and suffix ratio of 0.25, or last 25\% of each sequence as a suffix. DTO without KL achieves the highest and almost perfect forget quality of 0.9188 compared to an ideal retrained model, while most baselines remain below 0.055 and 0.16.  This is a significant margin, indicating DTO without KL is extremely effective in unlearning the target knowledge. DTO preserved model utility better by minimizing KL-divergence of logits on non-target tokens between unlearned model and original model. However, the forget quality is slightly lower. Moreover, DTO without KL is exhibiting the lowest Rouge-L score from the forget set (0.3725). This confirms that the actual response from the unlearned model is significantly less similar to the response of the forget set.
While both DTO and DTO without KL are showing model utility drop, the degradation from the original model is small, demonstrating a reasonable tradeoff given its strong forget quality.
FLAT exhibits better model utility, but their forget quality is low, indicating 
inherent knowledge about forget set still persists in the model. Both LLMU, DPO and NPO shows good model utility,

\begin{table*}[t] %
    \centering
    \scriptsize
    \begin{tabularx}{\textwidth}{l*{7}{>{\centering\arraybackslash}X}}
        \toprule
        &
        \multicolumn{3}{c}{\textbf{Llama 2-7B}} & \multicolumn{3}{c}{\textbf{Llama 3.2-1B}} \\
        \cmidrule(lr){2-4} \cmidrule(lr){5-7}
         & Forget Quality ($\uparrow$) & Model Utility ($\uparrow$) & Forget Rouge-L\,($\downarrow$) & Forget Quality ($\uparrow$) & Model Utility ($\uparrow$) & Forget Rouge-L\,($\downarrow$)\\
        \midrule
        Original LLM  & 4.513e-09 & 0.6319 & 0.8938 & 4.525e-08 & 0.6218 & 0.8250 \\
        Retrained LLM  & 1.0000  & 0.6263 & 0.3982 & 1.0000 & 0.6098 & 0.3857 \\
        \midrule
        LLMU          & 1.143e-05 & 0.3193 & 0.2310 & 0.0001 & 0.5928 & 0.6975 \\
        DPO           & 5.617e-06 & 0.4962 & 0.4857 & 0.0005 & 0.4966 & 0.4934 \\
        NPO           & 4.744e-06 & 0.5906 & 0.3977 & 0.0007 & 0.5536 & 0.4008 \\
        \midrule
        FLAT (TV)     & 0.0021 & 0.1253 & 0.0534 & 0.0124&  0.2071  & 0.0988\\
        FLAT (KL)     & 2.353e-05 & 0.2402 & 0.2832 & 0.0878 & 0.3266 & 0.1398 \\
        FLAT (JS)     & 0.0001 & 0.3091 & 0.1716 & 0.0001 & 0.4514 & 0.2118 \\
        FLAT (P)      & 1.873e-05 & 0.1971 & \textbf{0.0825} & 0.0030 & 0.2268 & \textbf{0.0902} \\
        \midrule
        \textbf{DTO w/o KL (Ours)} & 0.0021 & 0.2871 & 0.1743 & 0.0001 & 0.3187 & 0.2597 \\
        \textbf{DTO (Ours)} & \textbf{0.0876} & 0.4442 & 0.5415 & \textbf{0.3281} & 0.4218 & 0.3168 \\  
        \bottomrule
    \end{tabularx}
    \caption{Experimental results on TOFU 5\% dataset}%
    \label{tab:tofu05}
\end{table*}
\begin{table*}[h]
    \centering
    \scriptsize
    \begin{tabularx}{\textwidth}{l*{5}{>{\centering\arraybackslash}X}}
    \toprule
     & VerbMem on $D_u$ ($\downarrow$) & KnowMem on $D_u$ ($\downarrow$) & KnowMem on $D_r$ ($\uparrow$) & PrivLeak ($\downarrow$) \\
     \midrule
     Original Model &  99.70 & 45.87 & 68.40 & -58.19 \\
     \midrule
     LLMU & 99.70 & 44.60 & 67.69 & -57.37  \\
     DPO & 46.95 & 41.28 & 65.24 & -57.24 \\
     NPO & 68.85 & 30.65 & 48.96 & -53.90 \\
     \midrule 
     FLAT (TV) & 99.13 & 40.54 & 59.63 & -57.51 \\
     FLAT (KL) & 99.70 & 44.07 & 63.41 & -57.55 \\
     FLAT (JS) & \textbf{15.84} & 25.59 & 49.85 & -47.27 \\
     FLAT (P) & 98.22 & 43.00 & 60.89  & -57.59 \\
     \midrule
     \textbf{DTO w/o KL (Ours)} & 19.30 & \textbf{22.84} & 57.11 & \textbf{-47.14} \\
     \textbf{DTO (Ours)} & 88.42 & 46.21 & 66.40 & -57.22 \\
     \bottomrule
    \end{tabularx}
    \caption{Evaluation results on MUSE-Book dataset. Unlearn efficacy is assessed by VerbMem on $D_f$ ($\downarrow$), KnowMem on $D_f$ ($\downarrow$), and PrivLeak ($\downarrow$). Model utility is assessed by KnwoMem on $D_r$ ($\uparrow$).}
    \label{tab:muse_hp}
\end{table*}

Table~\ref{tab:tofu05} shows the result of unlearning 5\% of the TOFU dataset with DTO and baselines. LLMU, DPO and NPO have relatively high model utility and low forget quality. 
FLAT is showing both lower model utility and forget quality, implying over-unlearning. 
When the model loses linguistic capability due to harsh optimizations, it provides random answers for the QA dataset and shows different output distribution as the retrained model, resulting low forget quality. %
Moreover,
DTO shows better forget quality and better model utility than DTO without KL. By minimizing KL-divergence between original model over the logits of non-target tokens, the unlearned model was able to prevent losing linguistic capability, which leads to better forget quality. Compared to the baselines, DTO achieved best forget quality and relatively high model utility.
For results of unlearning 10\% TOFU data, please refer to Appendix~\ref{app:tofu_10}. Overall, DTO achieves the best forget quality over the TOFU dataset with comparable model utility.

Table~\ref{tab:muse_hp} shows the reuslt of unlearning MUSE-Books with DTO and baselines. LLMU, DPO and NPO failed to remove the verbatim memorization from forget set. Except for FLAT (JS), every other FLAT variants failed to remove the memorization. DTO w/o KL shows relatively small verbatim memorization. This was achievable because DTO directly suppresses tokens that trigger verbatim memorization. KnowMem on forget set evaluates more intrinsic knowledge memorization of forget set with QA datasets. DTO w/o KL achieved the lowest, showing that it also is capable of removing intrinsic knowledge. Similarly, KnowMem on retain data shows that DTO is able to keep the rest of the knowledge relatively intact. Lastly, PrivLeak shows the normalized membership inference risk compared to the retrained model. The negative sign means that the risk persists, and a score closer to zero means less risk. DTO w/o KL shows the closest score to zero among all baselines.

\begin{figure}[ht]
    \centering
    \begin{subfigure}[b]{0.5\textwidth}
        \includegraphics[width=\textwidth]{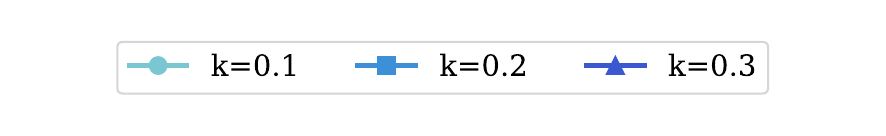}
    \end{subfigure}
    
    \vspace{-0.3cm}
    \begin{subfigure}[b]{0.49\textwidth}
        \includegraphics[width=\textwidth]{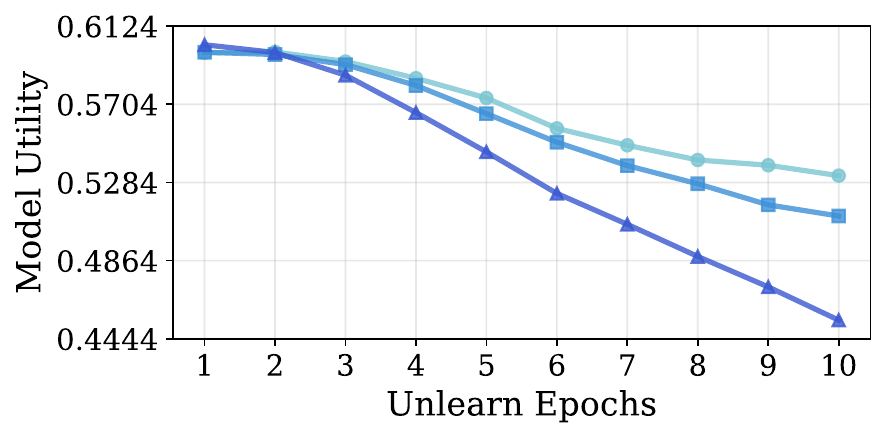}
    \end{subfigure} %
    \hfill
    \begin{subfigure}[b]{0.49\textwidth}
        \includegraphics[width=\textwidth]{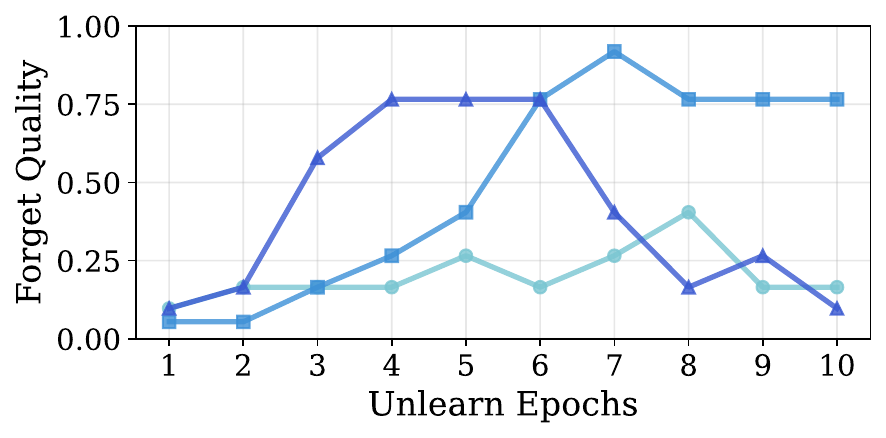}
    \end{subfigure}
    
    \caption{Model Utility and Forget quality with respect to various $k$. Suffix ratio is fixed to 0.25}
    \label{fig:k_abl}
\end{figure}

\begin{figure}[ht]
    \centering
    \begin{subfigure}[b]{0.6\textwidth}
        \includegraphics[width=\textwidth]{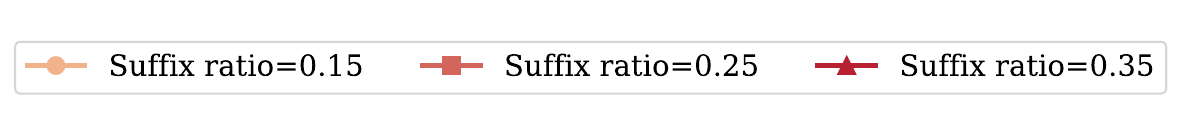}
    \end{subfigure}
     \vspace{-0.3cm}
    
    \begin{subfigure}[b]{0.49\textwidth}
        \includegraphics[width=\textwidth]{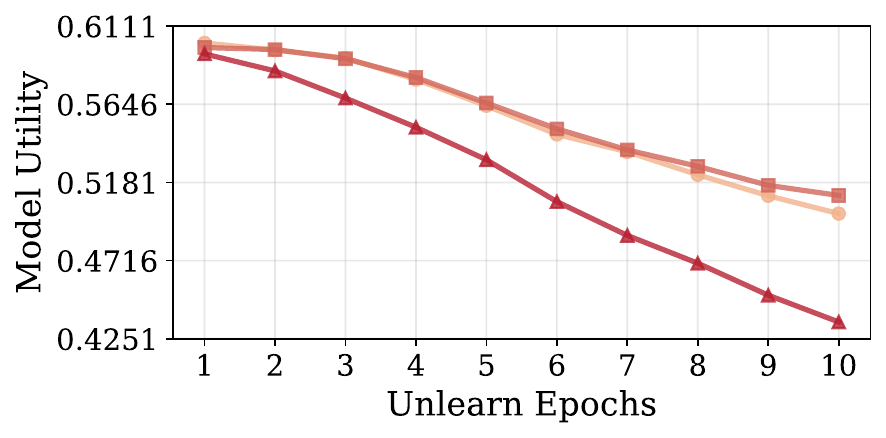}
    \end{subfigure}
    \hfill
    \begin{subfigure}[b]{0.49\textwidth}
        \includegraphics[width=\textwidth]{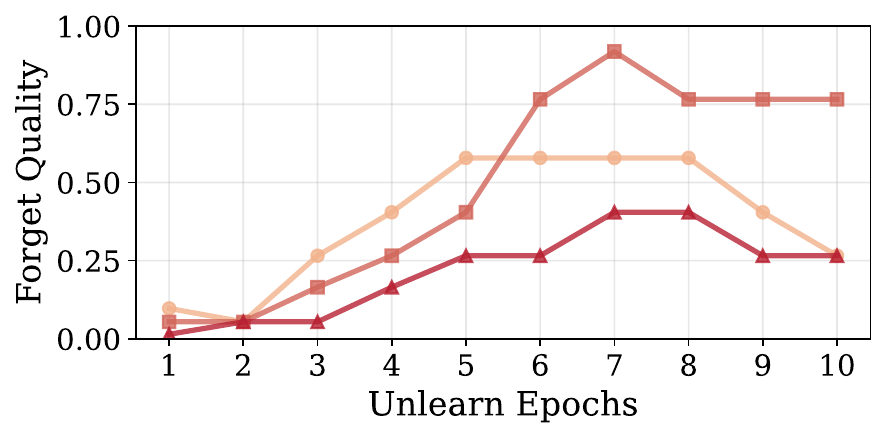}
    \end{subfigure}

    \caption{Model Utility and Forget quality with respect to various suffix ratios. $k$ is fixed to 0.2}
    \label{fig:suffix_abl}
\end{figure}

\partitle{Parameter Studies of Target Token Ratio and Suffix Ratio}
Figure~\ref{fig:k_abl} shows the progression of unlearning 1\% of TOFU dataset over various target token ratio $k$ with fixed suffix ratio. When $k=0.1$, top-10\% tokens with the highest delta-scores are selected as target tokens. Smaller size of the target tokens preserves model utility well, however, forget quality hardly increases, meaning that top-10\% were insufficient to unlearn. On the other hand, when $k=0.3$, the utility of the model drops rapidly and the quality of the forget increases quickly. However, after 6th epoch, forget quality starts to drop due to over-unlearning.

Figure~\ref{fig:suffix_abl} shows the progression of unlearning 1\% of TOFU dataset over various suffix ratios with fixed $k$. Delta-score for each prefix token is obtained from average NLL loss of suffix tokens. This makes the choice of suffix ratio critical. When suffix ratio is 0.15 (last 15\% of the tokens), model utility is largely preserved yet forget quality is less optimal. This is because a small suffix may contain only a limited portion of the target knowledge, providing insufficient signal for delta score to identify the most critical prefix tokens. Thus the selected target tokens from the prefix  might not fully  suppress the generation of target knowledge. Conversely, when suffix ratio is too large, irrelevant tokens start to influence the delta-score, introducing noise that leads to lower model utility and forget quality.
\begin{figure}[ht]
    \centering
    \begin{subfigure}[b]{0.49\textwidth}
        \includegraphics[width=\textwidth]{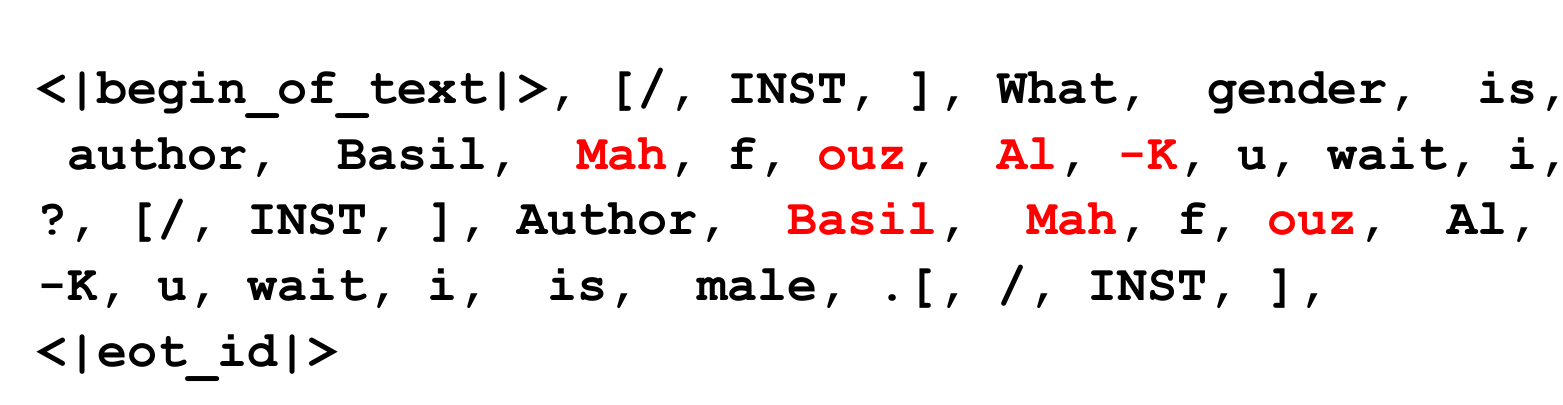}
        \caption{tokens selected by Delta-Score (Ours)}
        \label{fig:02_d}
    \end{subfigure}
    \hfill
    \begin{subfigure}[b]{0.49\textwidth}
        \includegraphics[width=\textwidth]{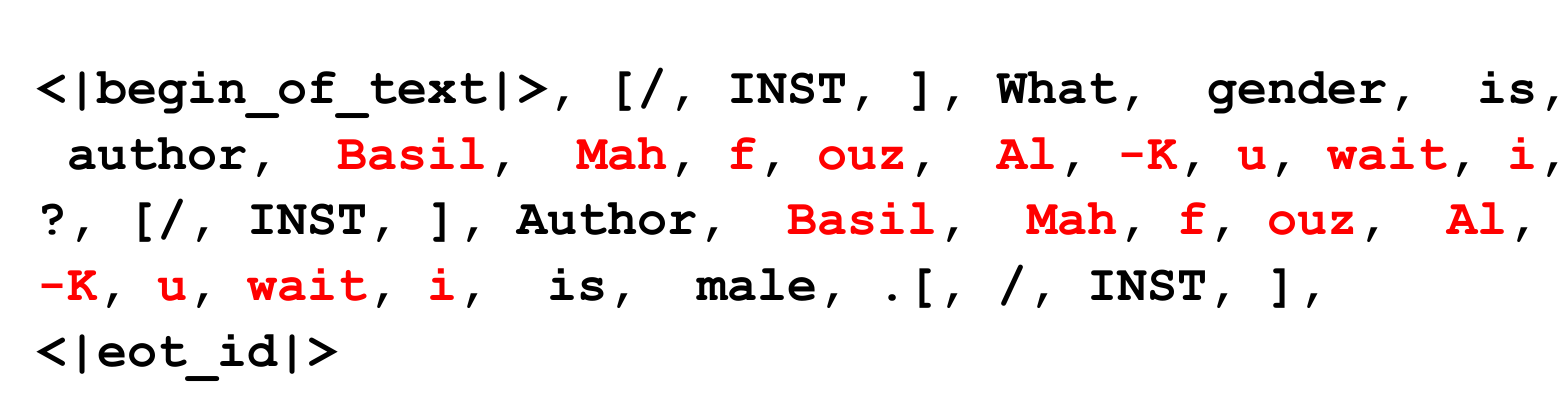}
        \caption{tokens selected by ChatGPT}
        \label{fig:02_c}
    \end{subfigure}
    \begin{subfigure}[b]{0.49\textwidth}
        \includegraphics[width=\textwidth]{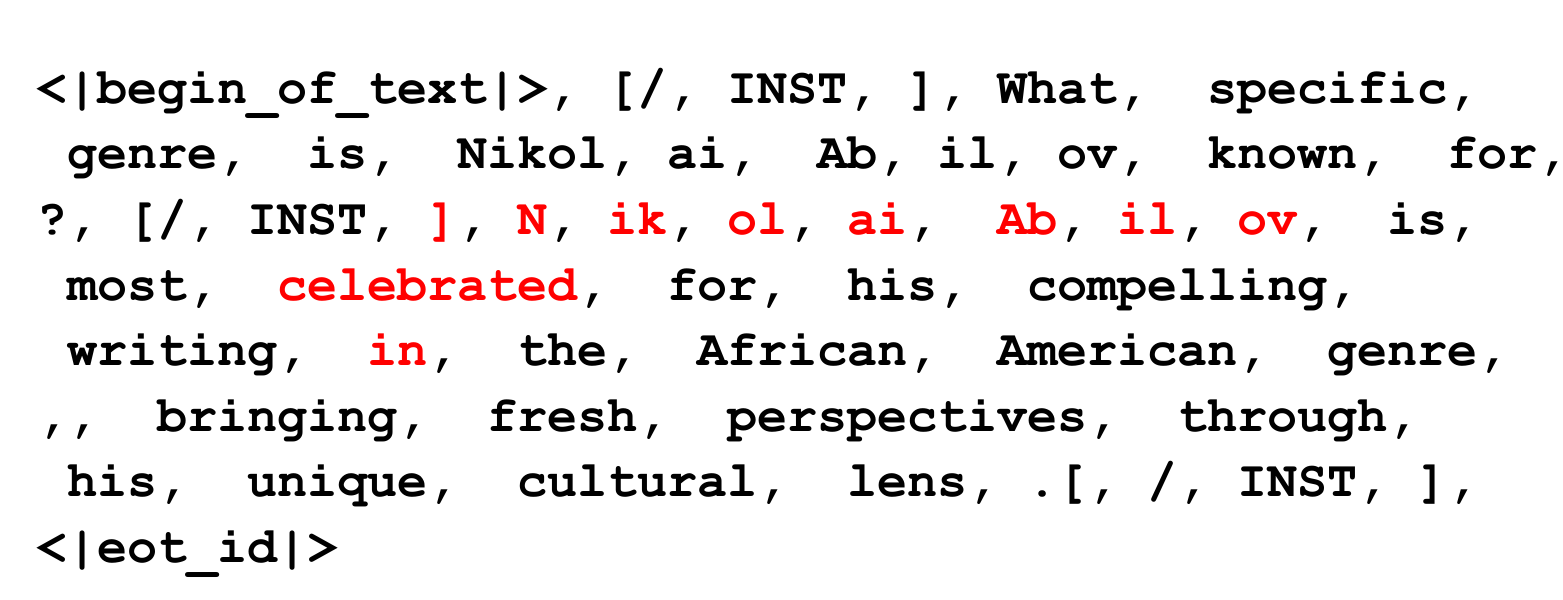}
        \caption{tokens selected from by Delta-Score (Ours)}
        \label{fig:26_d}
    \end{subfigure}
    \hfill
    \begin{subfigure}[b]{0.49\textwidth}
        \includegraphics[width=\textwidth]{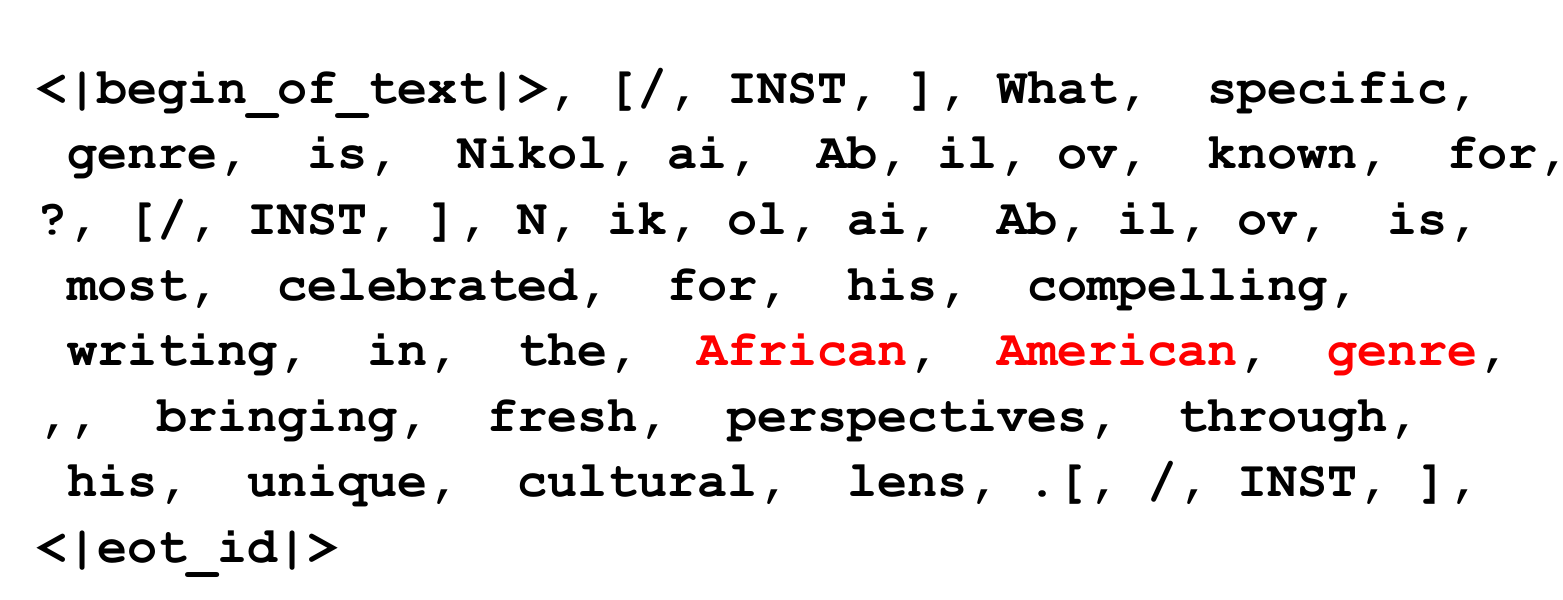}
        \caption{tokens selected from by ChatGPT}
        \label{fig:26_c}
    \end{subfigure}
    \caption{Selected tokens from sample 2 (first row) and 26 (second row) by the Delta-score and TPO}
    \label{fig:target_tokens}
\end{figure}
\begin{wraptable}[9]{r}{0.4\textwidth} %
    \centering
    \scriptsize
    \begin{tabularx}{\linewidth}{X|X|X}
        \toprule
             & Forget Quality & Model Utility \\
        \midrule
        TPO & 0.55 & 0.61 \\
        DTO & 0.26  & 0.42     \\
        \bottomrule
    \end{tabularx}
    \caption{Model Utility \% Forget Quality of unlearning 1\% of TOFU dataset from Llama 3.2-3b}
    \label{tab:tofu_1_3b}
\end{wraptable}

\partitle{Token Selection Strategy}
We compare our delta-score with the token selection strategy of Token Preference Optimization (TPO)~\citet{zhou2025not}, which uses ChatGPT to identify target \emph{words}, and use preference optimization to unlearn. We follow the instruction provided in the paper and prompt ChatGPT with the TOFU dataset. Figure~\ref{fig:target_tokens} shows the tokens selected by Delta-score and ChatGPT. Figure~\ref{fig:02_d} and~\ref{fig:02_c} shows that the delta-score primarily identified the last name of the person (Mahfouz Al-Kuwaiti) as targets similar to ChatGPT's selection, indicating that crucial tokens can be selected without external AI services. Figure~\ref{fig:26_d} and~\ref{fig:26_c} show that delta-score selected the name (Nikolai Abilov) and ``celerbrated" while ChatGPT picked the specific genre. The delta-score could not directly select the genre since it was part of the suffix. Instead, it selected the tokens that are directly relatable to the genre. 
Table~\ref{tab:tofu_1_3b} shows the result of unlearning 1\% of TOFU dataset from Llama 3.2-3B using DTO and TPO~\citep{zhou2025not}\footnote{We compare the result of TPO that is reported from the paper.}. 
Although DTO exhibits lower forget quality (which could be due to less than perfect target token selection and gradient ascent based unlearning), our token selection method avoids privacy risks associated with untrusted external AI services.
In addition, reducing the suffix length further improves DTO's selection correctness, as indicated by Figure~\ref{fig:suffix_abl}. Finally, our selection strategy is general and can use preference optimization instead of gradient ascent to enhance the unlearning. 

\section{Conclusion}
In this paper, we proposed Direct Token Optimization (DTO), a self-contained unlearning framework that unlearns an LLM without external resources, such as external reference models and retain datasets. Given a sequence to unlearn, DTO identifies target tokens that trigger the memorized knowledge using the proposed delta-score. During unlearning, target tokens are used for unlearning optimization while non-target tokens are used for maintaining model utility. Experimental results show that the DTO achieves substantially better forget quality than the state-of-the art methods while retaining reasonable model utility. In future work, we aim to incorporate preference optimization to improve the model utility forget quality trade-off, and further improve delta-score to more effectively select target tokens in each sequence.

\section*{Acknowledgment}
This research is supported by National Science Foundation under CNS-2437345, IIS-2302968, CNS-2124104, CNS-2125530, and National Institute of Health under R01ES033241 and R01LM013712. 

\section*{Reproducibility Statements}
We use the official dataset provided from TOFU~\citep{maini2024tofu} and MUSE~\citep{shi2024muse} benchmarks. For evaluations we use the official code provided from Open-Unlearning~\citep{dorna2025openunlearning}, a public gitub repository that provide comprehensive evaluation framework for these datasets. For unlearning TOFU data, we use the Llama-2-7B, Llama 3.2-1B, and Llama 3.2-3B models fine-tuned on the TOFU dataset, available in~\citet{dorna2025openunlearning} and ~\citet{maini2024tofu}. For MUSE, we use Llama-2-7B model officially fine-tuned on Chapter 2 of the Harry Potter series, available at~\citet{shi2024muse}.
We provide detailed hyperparameter settings in Appendix~\ref{app:param}. 

\section*{Ethics Statements}
Our proposed approach contributes to society and to human well-being by illustrating limitations and possible privacy risk of existing LLM unlearning methods, and proposing a novel machine unlearning method, which can protect individuals' data privacy in more rigorous manner. We strictly comply the code of ethics to ensure all data is properly handled, and conducted fair experiments.

\bibliography{iclr2026_conference}
\bibliographystyle{iclr2026_conference}

\newpage

\appendix
\section*{Appendix}

In this appendix session~\ref{app:llm} describes the use of large language models in this research. Section~\ref{app:param} provides detailed hyperparameter settings. Section~\ref{app:grad} illustrates gradient orthogonalization. Section~\ref{app:flat} discusses the integrity of our baseline implementation. Section~\ref{app:tofu_10} illustrates experimental results on unlearning TOFU 10\% dataset.

\section{LLM Usage}
\label{app:llm}

An LLM has been partially contributed to this research. LLMs assisted resolving minor technical issues on implementing baselines, our proposed method and experiments testbed. We rarely used LLM for assisting writing. While the LLM provided some writing suggestions, we did not directly copy and paste the LLM generated paragraphs into the paper.

\section{Hyperparmeter details and Implementations}
\label{app:param}
\begin{table*}[!h]
    \centering
    \small
    \begin{tabularx}{\textwidth}{*{6}{>{\centering\arraybackslash}X}}
        \toprule
        forget set & Model & $k$ & Suffix ratio & Batch size & Learning rate \\
        \midrule
        \multirow{2}{*}{TOFU 1\%} & Llama 3.2-1b & 0.2 & 0.25 & 8 & $1e-5$ \\
         & Llama 2-7b & 0.2 & 0.25 & 8 & $1e-5$ \\
         \midrule
         \multirow{2}{*}{TOFU 5\%} & Llama 3.2-1b & 0.2 & 0.2 & 8 & $2e-5$ \\
         & Llama 2-7b & 0.15 & 0.15 & 8 & $1e-5$ \\
         \midrule
         \multirow{2}{*}{TOFU 10\%} & Llama 3.2-1b & 0.1 & 0.1 & 8 & $5e-5$ \\
         & Llama 2-7b & 0.1 & 0.15 & 8 & $2e-5$ \\
         \midrule
         Muse-Books & Llama 2-7b & 0.2 & 0.25 & 8 & $1.5e-6$ \\
        \bottomrule  
    \end{tabularx}
    \caption{List of hyperparameters for DTO}
    \label{tab:hyperparameters}
\end{table*}

Table~\ref{tab:hyperparameters} shows the list of hyperparameters we used for each of the dataset. We used the same set of parameters for DTO and DTO without KL. %

\section{Gradient Orthogonalization}
\label{app:grad}
Gradient from unlearning loss and forget set and gradient from retain loss and retain set often has conflict. The directions often interfere themselves. Naively conducting each step-wise update leads to imperfect optimization for both objectives. This leads to a catastrophic utility loss.
Gradient orthogonalization reduces the gradient conflict by projecting a gradient~\citep{kodge2024deep}. Given two gradients $g_a$ and $g_b$, orthogonalizing $g_a$ to $g_b$ computs followings:
\begin{align}
    g^{orth}_a = g_b - \frac{\langle g_a, g_b \rangle}{\langle g_a, g_b \rangle}g_b.
\end{align}
Intuitively, this nullifies optimization directions in $g_a$ that are parallel to $g_b$, allowing less impact on the objective of $g_b$. In our method, we orthogonalize gradients from unlearn loss to the gradient of the retain loss, to achieve unlearn objective with less detrimental impact on model utility.

\section{Baseline implementations}
\label{app:flat}

\begin{table*}[!h] %
    \centering
    \small
    \begin{tabularx}{\textwidth}{l*{7}{>{\centering\arraybackslash}X}}
        \toprule
        &
        \multicolumn{3}{c}{\textbf{Our implementation}} & \multicolumn{3}{c}{\textbf{Reported in~\citet{wang2024llm}}} \\
        \cmidrule(lr){2-4} \cmidrule(lr){5-7}
         & Forget Quality ($\uparrow$) & Model Utility ($\uparrow$) & Forget Rouge-L\,($\downarrow$) & Forget Quality ($\uparrow$) & Model Utility ($\uparrow$) & Forget Rouge-L\,($\downarrow$)\\
        \midrule
        Original LLM  & 2.183e-06 & 0.6346 & 0.8849 & 4.488e-06 & 0.6346 & 0.9851 \\
        Retrained LLM  & 1.0000   & 0.6267 & 0.4080 & 1.0000 & 0.6267 & 0.0.4080 \\
        \midrule
        FLAT (TV)     & 0.0541 & 0.6199 & 0.4366 & 0.0541 & 0.6373 & 0.4391 \\
        FLAT (KL)     & 0.0301 & 0.6393 & 0.4971 & 0.0286 & 0.6393 & 0.5199 \\
        FLAT (JS)     & 0.0970 & 0.6214 & 0.4252 & 0.0541 & 0.6364 & 0.4454 \\
        FLAT (P)      & 0.0541 & 0.6239 & 0.4523 & 0.0541 & 0.6374 & 0.4392 \\
        \bottomrule
    \end{tabularx}
    \caption{Experimental result on TOFU 1\% dataset of FLAT implemented by us and the reported results.}
    \label{tab:flat_impl}
\end{table*}

DPO, NPO and LLMU have official implementations, however, FLAT~\citep{wang2024llm} is missing the implementation, hence we implemented them based on the paper. To verify integrity of our implementation, we compare our unlearning results with the reported results of FLAT. Table~\ref{tab:flat_impl} compares the result of unlearning 1\% of TOFU dataset on LLama 2-7B. Although model utility is slightly lower, the results show that our implementation is consistent with the reported results.

\section{Additional Experiments}
\label{app:tofu_10}

\begin{table*}[t] %
    \centering
    \scriptsize
    \begin{tabularx}{\textwidth}{l*{7}{>{\centering\arraybackslash}X}}
        \toprule
        &
        \multicolumn{3}{c}{\textbf{Llama 2-7B}} & \multicolumn{3}{c}{\textbf{Llama 3.2-1B}} \\
        \cmidrule(lr){2-4} \cmidrule(lr){5-7}
         & Forget Quality ($\uparrow$) & Model Utility ($\uparrow$) & Forget Rouge-L\,(↓) & Forget Quality ($\uparrow$) & Model Utility ($\uparrow$) & Forget Rouge-L\,(↓)\\
        \midrule
        Original LLM  & 1.735e-08 & 0.6346 & 0.8824 & 3.382e-06 & 0.6218 & 0.8194 \\
        Retrained LLM  & 1.0000  & 0.6122 & 0.3998 & 1.0000 & 0.5936 & 0.3785 \\
        \midrule
        LLMU          & 1.092e-06 & 0.2903 & 0.1127 & 4.353e-05 & 0.5795 & 0.6501 \\
        DPO           & 1.826e-07 & 0.5178 & 0.5745 & 1.119e-07 & 0.4764 & 0.4845 \\
        NPO           & 1.065e-06 & 0.5326 & 0.3587 & 1.839e-06 & 0.5429 & 0.4293 \\
        \midrule
        FLAT (TV)     & 4.35e-05 & 0.0866 & 0.0267 & 0.0013 & 0.2299 & 0.1367 \\
        FLAT (KL)     & 5.418e-05 & 0.0219 & 0.0013 & 0.0013 & 0.2727 & 0.1648 \\
        FLAT (JS)     & 4.587e-05 & 0.0802 & 0.0365 & 3.277e-05 & 0.3702 & 0.2952 \\
        FLAT (P)      &  0.0001 & 0.0538 & 0.0148  & 0.0005 & 0.1639 & 0.1089 \\
        \midrule
        \textbf{DTO w/o (Ours)} & 3.913e-06 & 0.1623 & 0.2118 & 4.353e-05 & 0.1971 & 0.2397 \\
        \textbf{DTO (Ours)} & \textbf{0.0004} & 0.4507 & 0.5580 & \textbf{0.0365} & 0.3067 & 0.3477 \\  
        \bottomrule
    \end{tabularx}
    \caption{Experimental results on TOFU 10\% dataset}%
    \label{tab:tofu10}
\end{table*}

Table~\ref{tab:tofu10} shows the result of unlearning 10\% of the TOFU dataset with DTO and baselines. For both models, LLMU, DPO and NPO failed to eliminate unlearn knowledge. FLAT achieved better forget quality, however, they suffer significaint utility loss. The utility loss is more significant from the 7B model than 1B model. We assume that $f$-divergence of FLAT over-generalizes the rejection template when the number of unlearning samples increases. DTO successfully reduced model utility loss, while achieving the best forget quality.

While DTO achieved the best forget quality among all baselines, forget quality failed to exceed 0.05, which serves as a statistical threshold for successful unlearning~\citep{mekala2024alternate}. Due to the size of the dataset, unlearning TOFU 10\% is challenging without the retain dataset. We aim to improve this in our future studies.

\end{document}

%% file: math_commands.tex
\usepackage{amsmath,amsfonts,bm}

\def\eqref#1{equation~\ref{#1}}

\def\1{\bm{1}}

\DeclareMathAlphabet{\mathsfit}{\encodingdefault}{\sfdefault}{m}{sl}
\SetMathAlphabet{\mathsfit}{bold}{\encodingdefault}{\sfdefault}{bx}{n}

%% file: iclr2026_conference.bbl
\begin{thebibliography}{61}
\providecommand{\natexlab}[1]{#1}
\providecommand{\url}[1]{\texttt{#1}}
\expandafter\ifx\csname urlstyle\endcsname\relax
  \providecommand{\doi}[1]{doi: #1}\else
  \providecommand{\doi}{doi: \begingroup \urlstyle{rm}\Url}\fi

\bibitem[Akkus et~al.(2025)Akkus, Aghdam, Li, Chu, Backes, Zhang, and Sav]{akkus2025generated}
Atilla Akkus, Masoud~Poorghaffar Aghdam, Mingjie Li, Junjie Chu, Michael Backes, Yuyang Zhang, and Sinem Sav.
\newblock Generated data with fake privacy: Hidden dangers of fine-tuning large language models on generated data.
\newblock In \emph{34th USENIX Security Symposium (USENIX Security 25)}, pp.\  8075--8093, 2025.

\bibitem[Basaran et~al.(2025)Basaran, Ahmed, Roy-Chowdhury, and Guler]{basaran2025a}
Umit~Yigit Basaran, Sk~Miraj Ahmed, Amit Roy-Chowdhury, and Basak Guler.
\newblock A certified unlearning approach without access to source data.
\newblock In \emph{Forty-second International Conference on Machine Learning}, 2025.
\newblock URL \url{https://openreview.net/forum?id=8lt5776GLB}.

\bibitem[Bourtoule et~al.(2021)Bourtoule, Chandrasekaran, Choquette-Choo, Jia, Travers, Zhang, Lie, and Papernot]{bourtoule2021machine}
Lucas Bourtoule, Varun Chandrasekaran, Christopher~A Choquette-Choo, Hengrui Jia, Adelin Travers, Baiwu Zhang, David Lie, and Nicolas Papernot.
\newblock Machine unlearning.
\newblock In \emph{2021 IEEE symposium on security and privacy (SP)}, pp.\  141--159. IEEE, 2021.

\bibitem[Cao \& Yang(2015)Cao and Yang]{cao2015towards}
Yinzhi Cao and Junfeng Yang.
\newblock Towards making systems forget with machine unlearning.
\newblock In \emph{2015 IEEE symposium on security and privacy}, pp.\  463--480. IEEE, 2015.

\bibitem[Carlini et~al.(2022{\natexlab{a}})Carlini, Chien, Nasr, Song, Terzis, and Tramer]{carlini2022membership}
Nicholas Carlini, Steve Chien, Milad Nasr, Shuang Song, Andreas Terzis, and Florian Tramer.
\newblock Membership inference attacks from first principles.
\newblock In \emph{2022 IEEE symposium on security and privacy (SP)}, pp.\  1897--1914. IEEE, 2022{\natexlab{a}}.

\bibitem[Carlini et~al.(2022{\natexlab{b}})Carlini, Ippolito, Jagielski, Lee, Tramer, and Zhang]{carlini2022quantifying}
Nicholas Carlini, Daphne Ippolito, Matthew Jagielski, Katherine Lee, Florian Tramer, and Chiyuan Zhang.
\newblock Quantifying memorization across neural language models.
\newblock In \emph{The Eleventh International Conference on Learning Representations}, 2022{\natexlab{b}}.

\bibitem[Cha et~al.(2024)Cha, Cho, Hwang, Lee, Moon, and Lee]{cha2024learning}
Sungmin Cha, Sungjun Cho, Dasol Hwang, Honglak Lee, Taesup Moon, and Moontae Lee.
\newblock Learning to unlearn: Instance-wise unlearning for pre-trained classifiers.
\newblock In \emph{Proceedings of the AAAI conference on artificial intelligence}, volume~38, pp.\  11186--11194, 2024.

\bibitem[Chen et~al.(2024)Chen, Ma, Zhang, Hao, Yan, Nourbakhsh, Yang, McAuley, Petzold, and Wang]{chen2024survey}
Zhiyu~Zoey Chen, Jing Ma, Xinlu Zhang, Nan Hao, An~Yan, Armineh Nourbakhsh, Xianjun Yang, Julian McAuley, Linda Petzold, and William~Yang Wang.
\newblock A survey on large language models for critical societal domains: Finance, healthcare, and law.
\newblock \emph{arXiv preprint arXiv:2405.01769}, 2024.

\bibitem[Chundawat et~al.(2023)Chundawat, Tarun, Mandal, and Kankanhalli]{chundawat2023zero}
Vikram~S Chundawat, Ayush~K Tarun, Murari Mandal, and Mohan Kankanhalli.
\newblock Zero-shot machine unlearning.
\newblock \emph{IEEE Transactions on Information Forensics and Security}, 18:\penalty0 2345--2354, 2023.

\bibitem[Cooper et~al.(2024)Cooper, Choquette-Choo, Bogen, Jagielski, Filippova, Liu, Chouldechova, Hayes, Huang, Mireshghallah, Shumailov, Triantafillou, Kairouz, Mitchell, Liang, Ho, Choi, Koyejo, Delgado, Grimmelmann, Shmatikov, Sa, Barocas, Cyphert, Lemley, boyd, Vaughan, Brundage, Bau, Neel, Jacobs, Terzis, Wallach, Papernot, and Lee]{cooper2024machine}
A.~Feder Cooper, Christopher~A. Choquette-Choo, Miranda Bogen, Matthew Jagielski, Katja Filippova, Ken~Ziyu Liu, Alexandra Chouldechova, Jamie Hayes, Yangsibo Huang, Niloofar Mireshghallah, Ilia Shumailov, Eleni Triantafillou, Peter Kairouz, Nicole Mitchell, Percy Liang, Daniel~E. Ho, Yejin Choi, Sanmi Koyejo, Fernando Delgado, James Grimmelmann, Vitaly Shmatikov, Christopher~De Sa, Solon Barocas, Amy Cyphert, Mark~A. Lemley, danah boyd, Jennifer~Wortman Vaughan, M.~Brundage, David Bau, Seth Neel, Abigail~Z. Jacobs, Andreas Terzis, Hanna Wallach, Nicolas Papernot, and Katherine Lee.
\newblock Machine unlearning doesn't do what you think: Lessons for generative ai policy, research, and practice.
\newblock Technical Report MSR-TR-2024-61, Microsoft, December 2024.

\bibitem[Deng et~al.(2025)Deng, Liu, Pang, He, Feng, Xuan, Zhu, and Wei]{deng2025guard}
Zhijie Deng, Chris~Yuhao Liu, Zirui Pang, Xinlei He, Lei Feng, Qi~Xuan, Zhaowei Zhu, and Jiaheng Wei.
\newblock Guard: Generation-time llm unlearning via adaptive restriction and detection.
\newblock \emph{arXiv preprint arXiv:2505.13312}, 2025.

\bibitem[Dorna et~al.(2025)Dorna, Mekala, Zhao, McCallum, Lipton, Kolter, and Maini]{dorna2025openunlearning}
Vineeth Dorna, Anmol Mekala, Wenlong Zhao, Andrew McCallum, Zachary~C Lipton, J~Zico Kolter, and Pratyush Maini.
\newblock Openunlearning: Accelerating llm unlearning via unified benchmarking of methods and metrics.
\newblock \emph{arXiv preprint arXiv:2506.12618}, 2025.

\bibitem[Duan et~al.(2024)Duan, Suri, Mireshghallah, Min, Shi, Zettlemoyer, Tsvetkov, Choi, Evans, and Hajishirzi]{duan2024membership}
Michael Duan, Anshuman Suri, Niloofar Mireshghallah, Sewon Min, Weijia Shi, Luke Zettlemoyer, Yulia Tsvetkov, Yejin Choi, David Evans, and Hannaneh Hajishirzi.
\newblock Do membership inference attacks work on large language models?
\newblock \emph{arXiv preprint arXiv:2402.07841}, 2024.

\bibitem[Ebrahimpour-Boroojeny et~al.(2025)Ebrahimpour-Boroojeny, Sundaram, and Chandrasekaran]{ebrahimpournot}
Ali Ebrahimpour-Boroojeny, Hari Sundaram, and Varun Chandrasekaran.
\newblock Not all wrong is bad: Using adversarial examples for unlearning.
\newblock In \emph{Forty-second International Conference on Machine Learning (ICML 2025)}, 2025.

\bibitem[Eldan \& Russinovich(2023)Eldan and Russinovich]{eldan2023s}
Ronen Eldan and Mark Russinovich.
\newblock Who’s harry potter? approximate unlearning for llms.
\newblock 2023.

\bibitem[Fan et~al.(2024)Fan, Liu, Lin, Jia, Zhang, Mei, and Liu]{fan2024simplicity}
Chongyu Fan, Jiancheng Liu, Licong Lin, Jinghan Jia, Ruiqi Zhang, Song Mei, and Sijia Liu.
\newblock Simplicity prevails: Rethinking negative preference optimization for llm unlearning.
\newblock \emph{arXiv preprint arXiv:2410.07163}, 2024.

\bibitem[Fu et~al.(2024)Fu, Wang, Gao, Liu, Li, and Jiang]{fu2024membership}
Wenjie Fu, Huandong Wang, Chen Gao, Guanghua Liu, Yong Li, and Tao Jiang.
\newblock Membership inference attacks against fine-tuned large language models via self-prompt calibration.
\newblock \emph{Advances in Neural Information Processing Systems}, 37:\penalty0 134981--135010, 2024.

\bibitem[Gao et~al.(2024)Gao, Wang, Ding, Weng, Wang, and Zhu]{gao2024large}
Chongyang Gao, Lixu Wang, Kaize Ding, Chenkai Weng, Xiao Wang, and Qi~Zhu.
\newblock On large language model continual unlearning.
\newblock \emph{arXiv preprint arXiv:2407.10223}, 2024.

\bibitem[Gao et~al.(2020)Gao, Biderman, Black, Golding, Hoppe, Foster, Phang, He, Thite, Nabeshima, et~al.]{gao2020pile}
Leo Gao, Stella Biderman, Sid Black, Laurence Golding, Travis Hoppe, Charles Foster, Jason Phang, Horace He, Anish Thite, Noa Nabeshima, et~al.
\newblock The pile: An 800gb dataset of diverse text for language modeling.
\newblock \emph{arXiv preprint arXiv:2101.00027}, 2020.

\bibitem[Gu et~al.(2024)Gu, Rashid, Sultana, and Mehnaz]{gu2024second}
Kang Gu, Md~Rafi~Ur Rashid, Najrin Sultana, and Shagufta Mehnaz.
\newblock Second-order information matters: Revisiting machine unlearning for large language models.
\newblock \emph{arXiv preprint arXiv:2403.10557}, 2024.

\bibitem[Guo et~al.(2019)Guo, Goldstein, Hannun, and Van Der~Maaten]{guo2019certified}
Chuan Guo, Tom Goldstein, Awni Hannun, and Laurens Van Der~Maaten.
\newblock Certified data removal from machine learning models.
\newblock \emph{arXiv preprint arXiv:1911.03030}, 2019.

\bibitem[Huang et~al.(2024{\natexlab{a}})Huang, Yang, and Potts]{huang2024demystifying}
Jing Huang, Diyi Yang, and Christopher Potts.
\newblock Demystifying verbatim memorization in large language models.
\newblock \emph{arXiv preprint arXiv:2407.17817}, 2024{\natexlab{a}}.

\bibitem[Huang et~al.(2024{\natexlab{b}})Huang, Cheng, Zheng, Wang, He, Li, and Huang]{huang2024unified}
Zhehao Huang, Xinwen Cheng, JingHao Zheng, Haoran Wang, Zhengbao He, Tao Li, and Xiaolin Huang.
\newblock Unified gradient-based machine unlearning with remain geometry enhancement.
\newblock \emph{Advances in Neural Information Processing Systems}, 37:\penalty0 26377--26414, 2024{\natexlab{b}}.

\bibitem[Ji et~al.(2024)Ji, Liu, Zhang, Liu, Kompella, Liu, and Chang]{ji2024reversing}
Jiabao Ji, Yujian Liu, Yang Zhang, Gaowen Liu, Ramana~R Kompella, Sijia Liu, and Shiyu Chang.
\newblock Reversing the forget-retain objectives: An efficient llm unlearning framework from logit difference.
\newblock \emph{Advances in Neural Information Processing Systems}, 37:\penalty0 12581--12611, 2024.

\bibitem[Jia et~al.(2024)Jia, Zhang, Zhang, Liu, Runwal, Diffenderfer, Kailkhura, and Liu]{jia2024soul}
Jinghan Jia, Yihua Zhang, Yimeng Zhang, Jiancheng Liu, Bharat Runwal, James Diffenderfer, Bhavya Kailkhura, and Sijia Liu.
\newblock Soul: Unlocking the power of second-order optimization for llm unlearning.
\newblock \emph{arXiv preprint arXiv:2404.18239}, 2024.

\bibitem[Jin et~al.(2024)Jin, Cao, Wang, He, Yuan, Li, Chen, Liu, and Zhao]{jin2024rwku}
Zhuoran Jin, Pengfei Cao, Chenhao Wang, Zhitao He, Hongbang Yuan, Jiachun Li, Yubo Chen, Kang Liu, and Jun Zhao.
\newblock Rwku: Benchmarking real-world knowledge unlearning for large language models.
\newblock \emph{Advances in Neural Information Processing Systems}, 37:\penalty0 98213--98263, 2024.

\bibitem[Kodge et~al.(2024)Kodge, Saha, and Roy]{kodge2024deep}
Sangamesh Kodge, Gobinda Saha, and Kaushik Roy.
\newblock Deep unlearning: Fast and efficient gradient-free class forgetting.
\newblock \emph{Transactions on Machine Learning Research}, 2024.

\bibitem[Koloskova et~al.(2025)Koloskova, Allouah, Jha, Guerraoui, and Koyejo]{koloskova2025certified}
Anastasia Koloskova, Youssef Allouah, Animesh Jha, Rachid Guerraoui, and Sanmi Koyejo.
\newblock Certified unlearning for neural networks, 2025.

\bibitem[Krishnan et~al.(2025)Krishnan, Reddy, and Mosbach]{krishnan2025not}
Aravind Krishnan, Siva Reddy, and Marius Mosbach.
\newblock Not all data are unlearned equally.
\newblock \emph{arXiv preprint arXiv:2504.05058}, 2025.

\bibitem[Kumar(2025)]{kumar2025selective}
Varun~Sampath Kumar.
\newblock Selective knowledge unlearning via self-distillation with auxiliary forget-set model.
\newblock In \emph{ICML 2025 Workshop on Machine Unlearning for Generative AI}, 2025.

\bibitem[Kurmanji et~al.(2023)Kurmanji, Triantafillou, Hayes, and Triantafillou]{kurmanji2023towards}
Meghdad Kurmanji, Peter Triantafillou, Jamie Hayes, and Eleni Triantafillou.
\newblock Towards unbounded machine unlearning.
\newblock \emph{Advances in neural information processing systems}, 36:\penalty0 1957--1987, 2023.

\bibitem[Lee et~al.(2025)Lee, Zhang, Yang, Lou, and Xiong]{lee2024contrastive}
Hong~kyu Lee, Qiuchen Zhang, Carl Yang, Jian Lou, and Li~Xiong.
\newblock Contrastive unlearning: A contrastive approach to machine unlearning.
\newblock In \emph{the 34th International Joint Conference on Artificial Intelligence (IJCAI 2025)}, 2025.

\bibitem[Li et~al.(2024)Li, Pan, Gopal, Yue, Berrios, Gatti, Li, Dombrowski, Goel, Phan, et~al.]{li2024wmdp}
Nathaniel Li, Alexander Pan, Anjali Gopal, Summer Yue, Daniel Berrios, Alice Gatti, Justin~D Li, Ann-Kathrin Dombrowski, Shashwat Goel, Long Phan, et~al.
\newblock The wmdp benchmark: Measuring and reducing malicious use with unlearning.
\newblock \emph{arXiv preprint arXiv:2403.03218}, 2024.

\bibitem[Liu et~al.(2024{\natexlab{a}})Liu, Wang, Flanigan, and Liu]{liu2024large}
Chris Liu, Yaxuan Wang, Jeffrey Flanigan, and Yang Liu.
\newblock Large language model unlearning via embedding-corrupted prompts.
\newblock \emph{Advances in Neural Information Processing Systems}, 37:\penalty0 118198--118266, 2024{\natexlab{a}}.

\bibitem[Liu et~al.(2024{\natexlab{b}})Liu, Zhang, Jaakkola, and Chang]{liu2024revisiting}
Yujian Liu, Yang Zhang, Tommi Jaakkola, and Shiyu Chang.
\newblock Revisiting who's harry potter: Towards targeted unlearning from a causal intervention perspective.
\newblock \emph{arXiv preprint arXiv:2407.16997}, 2024{\natexlab{b}}.

\bibitem[Liu et~al.(2024{\natexlab{c}})Liu, Dou, Tan, Tian, and Jiang]{liu2024towards}
Zheyuan Liu, Guangyao Dou, Zhaoxuan Tan, Yijun Tian, and Meng Jiang.
\newblock Towards safer large language models through machine unlearning.
\newblock \emph{arXiv preprint arXiv:2402.10058}, 2024{\natexlab{c}}.

\bibitem[Maini et~al.(2024)Maini, Feng, Schwarzschild, Lipton, and Kolter]{maini2024tofu}
Pratyush Maini, Zhili Feng, Avi Schwarzschild, Zachary~C Lipton, and J~Zico Kolter.
\newblock Tofu: A task of fictitious unlearning for llms.
\newblock \emph{arXiv preprint arXiv:2401.06121}, 2024.

\bibitem[Mantelero(2013)]{mantelero_eu_2013}
Alessandro Mantelero.
\newblock The {EU} proposal for a general data protection regulation and the roots of the ‘right to be forgotten’.
\newblock \emph{Computer Law \& Security Review}, 29\penalty0 (3):\penalty0 229--235, 2013.
\newblock ISSN 0267-3649.
\newblock \doi{10.1016/j.clsr.2013.03.010}.
\newblock URL \url{https://www.sciencedirect.com/science/article/pii/S0267364913000654}.

\bibitem[Mekala et~al.(2024)Mekala, Dorna, Dubey, Lalwani, Koleczek, Rungta, Hasan, and Lobo]{mekala2024alternate}
Anmol Mekala, Vineeth Dorna, Shreya Dubey, Abhishek Lalwani, David Koleczek, Mukund Rungta, Sadid Hasan, and Elita Lobo.
\newblock Alternate preference optimization for unlearning factual knowledge in large language models.
\newblock \emph{arXiv preprint arXiv:2409.13474}, 2024.

\bibitem[{OpenAI}(2025)]{openai_privacy_policy_2025}
{OpenAI}.
\newblock Openai privacy policy, 2025.
\newblock URL \url{https://openai.com/policies/privacy-policy}.
\newblock Version updated June 27, 2025.

\bibitem[Rafailov et~al.(2023)Rafailov, Sharma, Mitchell, Manning, Ermon, and Finn]{rafailov2023direct}
Rafael Rafailov, Archit Sharma, Eric Mitchell, Christopher~D Manning, Stefano Ermon, and Chelsea Finn.
\newblock Direct preference optimization: Your language model is secretly a reward model.
\newblock \emph{Advances in neural information processing systems}, 36:\penalty0 53728--53741, 2023.

\bibitem[Shi et~al.(2024{\natexlab{a}})Shi, Tan, Qiu, Qu, Nie, Cheng, Chu, Yinghui, and Qi]{shi2024ulmr}
Shaojie Shi, Xiaoyu Tan, Xihe Qiu, Chao Qu, Kexin Nie, Yuan Cheng, Wei Chu, Xu~Yinghui, and Yuan Qi.
\newblock Ulmr: Unlearning large language models via negative response and model parameter average.
\newblock In \emph{Proceedings of the 2024 Conference on Empirical Methods in Natural Language Processing: Industry Track}, pp.\  755--762, 2024{\natexlab{a}}.

\bibitem[Shi et~al.(2024{\natexlab{b}})Shi, Lee, Huang, Malladi, Zhao, Holtzman, Liu, Zettlemoyer, Smith, and Zhang]{shi2024muse}
Weijia Shi, Jaechan Lee, Yangsibo Huang, Sadhika Malladi, Jieyu Zhao, Ari Holtzman, Daogao Liu, Luke Zettlemoyer, Noah~A Smith, and Chiyuan Zhang.
\newblock Muse: Machine unlearning six-way evaluation for language models.
\newblock \emph{arXiv preprint arXiv:2407.06460}, 2024{\natexlab{b}}.

\bibitem[Stoehr et~al.(2024)Stoehr, Gordon, Zhang, and Lewis]{stoehr2024localizing}
Niklas Stoehr, Mitchell Gordon, Chiyuan Zhang, and Owen Lewis.
\newblock Localizing paragraph memorization in language models.
\newblock \emph{arXiv preprint arXiv:2403.19851}, 2024.

\bibitem[Tarun et~al.(2023)Tarun, Chundawat, Mandal, and Kankanhalli]{tarun2023fast}
Ayush~K Tarun, Vikram~S Chundawat, Murari Mandal, and Mohan Kankanhalli.
\newblock Fast yet effective machine unlearning.
\newblock \emph{IEEE Transactions on Neural Networks and Learning Systems}, 35\penalty0 (9):\penalty0 13046--13055, 2023.

\bibitem[Tirumala et~al.(2022)Tirumala, Markosyan, Zettlemoyer, and Aghajanyan]{tirumala2022memorization}
Kushal Tirumala, Aram Markosyan, Luke Zettlemoyer, and Armen Aghajanyan.
\newblock Memorization without overfitting: Analyzing the training dynamics of large language models.
\newblock \emph{Advances in Neural Information Processing Systems}, 35:\penalty0 38274--38290, 2022.

\bibitem[Touvron et~al.(2023)Touvron, Martin, Stone, Albert, Almahairi, Babaei, Bashlykov, Batra, Bhargava, Bhosale, et~al.]{touvron2023llama}
Hugo Touvron, Louis Martin, Kevin Stone, Peter Albert, Amjad Almahairi, Yasmine Babaei, Nikolay Bashlykov, Soumya Batra, Prajjwal Bhargava, Shruti Bhosale, et~al.
\newblock Llama 2: Open foundation and fine-tuned chat models.
\newblock \emph{arXiv preprint arXiv:2307.09288}, 2023.

\bibitem[Wang et~al.(2025{\natexlab{a}})Wang, Zhou, Zhou, Shin, Han, and Weinberger]{wang2025rethinking}
Qizhou Wang, Jin~Peng Zhou, Zhanke Zhou, Saebyeol Shin, Bo~Han, and Kilian~Q Weinberger.
\newblock Rethinking {LLM} unlearning objectives: A gradient perspective and go beyond.
\newblock In \emph{The Thirteenth International Conference on Learning Representations}, 2025{\natexlab{a}}.
\newblock URL \url{https://openreview.net/forum?id=huo8MqVH6t}.

\bibitem[Wang et~al.(2024)Wang, Zhu, Liu, Ding, Guo, Ye, Zhou, and Yu]{wang2024unique}
Shang Wang, Tianqing Zhu, Bo~Liu, Ming Ding, Xu~Guo, Dayong Ye, Wanlei Zhou, and Philip~S Yu.
\newblock Unique security and privacy threats of large language model: A comprehensive survey.
\newblock \emph{arXiv preprint arXiv:2406.07973}, 2024.

\bibitem[Wang et~al.(2025{\natexlab{b}})Wang, Wei, Liu, Pang, Liu, Shah, Bao, Liu, and Wei]{wang2024llm}
Yaxuan Wang, Jiaheng Wei, Chris~Yuhao Liu, Jinlong Pang, Quan Liu, Ankit~Parag Shah, Yujia Bao, Yang Liu, and Wei Wei.
\newblock Llm unlearning via loss adjustment with only forget data.
\newblock In \emph{The Thirteenth International Conference on Learning Representations}, 2025{\natexlab{b}}.

\bibitem[Wu et~al.(2024)Wu, Duan, and Ni]{wu2024unveiling}
Xiaodong Wu, Ran Duan, and Jianbing Ni.
\newblock Unveiling security, privacy, and ethical concerns of chatgpt.
\newblock \emph{Journal of information and intelligence}, 2\penalty0 (2):\penalty0 102--115, 2024.

\bibitem[Xiao et~al.(2025)Xiao, Zhou, Liu, Liu, Li, Liu, and Huang]{xiao2025comprehensive}
Hanguang Xiao, Feizhong Zhou, Xingyue Liu, Tianqi Liu, Zhipeng Li, Xin Liu, and Xiaoxuan Huang.
\newblock A comprehensive survey of large language models and multimodal large language models in medicine.
\newblock \emph{Information Fusion}, 117:\penalty0 102888, 2025.

\bibitem[Yang et~al.(2025)Yang, Wang, Huang, Liu, Zhang, and Han]{yang2025exploring}
Puning Yang, Qizhou Wang, Zhuo Huang, Tongliang Liu, Chengqi Zhang, and Bo~Han.
\newblock Exploring criteria of loss reweighting to enhance llm unlearning.
\newblock In \emph{Proceedings of the 42nd International Conference on Machine Learning (ICML)}, 2025.

\bibitem[Yao et~al.(2024{\natexlab{a}})Yao, Chien, Du, Niu, Wang, Cheng, and Yue]{yao2024machine}
Jin Yao, Eli Chien, Minxin Du, Xinyao Niu, Tianhao Wang, Zezhou Cheng, and Xiang Yue.
\newblock Machine unlearning of pre-trained large language models.
\newblock \emph{arXiv preprint arXiv:2402.15159}, 2024{\natexlab{a}}.

\bibitem[Yao et~al.(2024{\natexlab{b}})Yao, Xu, and Liu]{yao2024large}
Yuanshun Yao, Xiaojun Xu, and Yang Liu.
\newblock Large language model unlearning.
\newblock \emph{Advances in Neural Information Processing Systems}, 37:\penalty0 105425--105475, 2024{\natexlab{b}}.

\bibitem[Yuan et~al.(2024)Yuan, Pang, Du, Chen, Zhang, and Lin]{yuan2024closer}
Xiaojian Yuan, Tianyu Pang, Chao Du, Kejiang Chen, Weiming Zhang, and Min Lin.
\newblock A closer look at machine unlearning for large language models.
\newblock \emph{arXiv preprint arXiv:2410.08109}, 2024.

\bibitem[Zeng et~al.(2023)Zeng, Li, Ren, Liu, Xu, He, Xing, Wang, Tang, and Yin]{zeng2023exploring}
Shenglai Zeng, Yaxin Li, Jie Ren, Yiding Liu, Han Xu, Pengfei He, Yue Xing, Shuaiqiang Wang, Jiliang Tang, and Dawei Yin.
\newblock Exploring memorization in fine-tuned language models.
\newblock \emph{arXiv preprint arXiv:2310.06714}, 2023.

\bibitem[Zhang et~al.(2024)Zhang, Lin, Bai, and Mei]{zhang2024negative}
Ruiqi Zhang, Licong Lin, Yu~Bai, and Song Mei.
\newblock Negative preference optimization: From catastrophic collapse to effective unlearning.
\newblock \emph{arXiv preprint arXiv:2404.05868}, 2024.

\bibitem[Zhou et~al.(2024)Zhou, Wang, Ye, Wu, and Chang]{zhou2024limitations}
Shiji Zhou, Lianzhe Wang, Jiangnan Ye, Yongliang Wu, and Heng Chang.
\newblock On the limitations and prospects of machine unlearning for generative ai.
\newblock \emph{arXiv preprint arXiv:2408.00376}, 2024.

\bibitem[Zhou et~al.(2025)Zhou, Qiang, Zade, Zytko, Khanduri, and Zhu]{zhou2025not}
Xiangyu Zhou, Yao Qiang, Saleh~Zare Zade, Douglas Zytko, Prashant Khanduri, and Dongxiao Zhu.
\newblock Not all tokens are meant to be forgotten.
\newblock \emph{arXiv preprint arXiv:2506.03142}, 2025.

\bibitem[Ziegler et~al.(2019)Ziegler, Stiennon, Wu, Brown, Radford, Amodei, Christiano, and Irving]{ziegler2019fine}
Daniel~M Ziegler, Nisan Stiennon, Jeffrey Wu, Tom~B Brown, Alec Radford, Dario Amodei, Paul Christiano, and Geoffrey Irving.
\newblock Fine-tuning language models from human preferences.
\newblock \emph{arXiv preprint arXiv:1909.08593}, 2019.

\end{thebibliography}
